\documentclass[letterpaper, 10pt, conference]{ieeeconf}  

\IEEEoverridecommandlockouts                              

\usepackage{cite}
\usepackage{graphicx}
\usepackage{amsmath}
\usepackage{amssymb}
\usepackage{url}
\usepackage{enumerate}
\usepackage{booktabs}
\usepackage{multirow}
\usepackage{xcolor}

\title{\LARGE \bf
Profit-Based Counterfactual Explanations for Product Improvement:\\
A Case Study of Manga Sales in Japan*%
}

\author{Keita Kinjo$^{1}$ and Takeshi Ebina$^{2}$%
\thanks{* Research supported by JSPS KAKENHI Grant-in-Aid for Scientific Research.}%
\thanks{$^{1}$K. Kinjo is with the Faculty of Business Studies, Kyoritsu Women's
        University, 2-2-1, Hitotsubashi, Chiyoda-ku, Tokyo, 101-8437, Japan
        (corresponding author; phone: +81-3-3237-2159;
        {\tt\small kkinjo@kyoritsu-wu.ac.jp})}%
\thanks{$^{2}$T. Ebina is with the School of Commerce, Meiji University,
        1-1, Kanda-surugadai, Chiyoda-ku, Tokyo, 101-8301, Japan
        {\tt\small ebina@meiji.ac.jp}}%
}

\begin{document}

\maketitle
\thispagestyle{empty}
\pagestyle{empty}

\begin{abstract}

Counterfactual explanation (CE) is widely used to enhance the
interpretability of machine learning models and support data-driven
decision-making based on model predictions. However, existing CE methods
typically require two exogenously specified inputs: a desired output value
(target) and a distance function that quantifies changes in explanatory
variables. In regression settings, neither the validity of target
specification nor the practical interpretation of the distance metric has
been sufficiently addressed. Furthermore, most existing CE methods focus on
altering predictions rather than optimizing a decision objective, even though
real-world decision-making often requires explicit objective maximization. To
address these limitations, we formulate CE as a profit maximization problem
in management and marketing contexts and propose a framework termed
profit-based counterfactual explanation (PBCE). PBCE eliminates the need for
exogenous target specification by directly maximizing profit as the primary
optimization objective. Concurrently, the distance term is reinterpreted as
the cost of modifying product attributes, providing a clear and economically
grounded interpretation.

\end{abstract}

\section{INTRODUCTION}

Applications of artificial intelligence (AI) and machine learning have been
rapidly expanding across a wide range of fields. In particular,
machine-learning-based prediction and control methods are increasingly
utilized in domains such as economics, healthcare, and marketing. Despite
their strong predictive performance, many machine learning models possess
highly complex internal structures and are often regarded as black-box models,
in which the reasoning behind the predictions is not explicitly interpretable.
Consequently, understanding why a particular prediction was made, or how
decision-makers should respond to it, is often difficult.

Therefore, considerable research has been focused on improving the
interpretability and explainability of machine learning models \cite{c1}. Many
approaches have been proposed for extracting human-interpretable information
from black-box models. Counterfactual explanations (CE) have emerged as a
prominent framework for explaining individual predictions \cite{c2,c3,c4}. The basic
concept of CE is as follows: Given a trained machine learning model and the
predicted value for a specific instance, the CE identifies how explanatory
variables should be modified to move the predicted outcome toward a desired
target value. Simultaneously, a constraint is typically imposed to ensure
that the modified explanatory variables do not deviate excessively from their
original input values. This is usually implemented by minimizing the distance
between the original and counterfactual inputs. In other words, the CE problem
is formulated as finding counterfactual instances that jointly minimize
prediction loss with respect to the target value and distance from the
original input. By presenting counterfactual examples, CE enables users to
identify explanatory variables that play an important role in generating
predictions. The concept of CE is also closely related to algorithmic recourse
and has theoretical connections with research on adversarial example
generation \cite{c4}. A wide range of CE methods have been proposed, differing in
model type (e.g., differentiable vs.\ non-differentiable models), extraction
strategy, data format (e.g., tabular, image, or text data), and design
objectives such as actionability, plausibility, and fairness \cite{c4,c5}.

Despite these technological advances, several important challenges remain for
the CE framework. The first is the issue of specifying a target value
exogenously. In classification problems, a natural objective is defined by
flipping a predicted label. In contrast, in regression settings, the
appropriate target value is not clearly defined. Although many studies have
assumed that the target value is given exogenously, the practical validity of
this assumption has not yet been sufficiently discussed. Second, there is the
issue of interpreting distance functions. In CE, minimizing the distance
between the original and counterfactual input is common \cite{c6}, as this is
assumed to improve the acceptability and feasibility of the explanation \cite{c7}.
However, the distance function is typically defined exogenously, for example,
using the Euclidean distance, and its correspondence with the real-world cost
of modifying explanatory variables is often unclear. Third, most existing CE
studies focus primarily on how predictions can be altered rather than on what
decisions should be optimized. In practical decision-making contexts, defining
an appropriate objective function and optimizing it is often more important
than merely moving predictions toward a particular target value.

Motivated by these issues, this study addresses the following research
question: How can CEs be formulated to directly support profit-oriented
managerial decision-making? To answer this question, focusing on product
strategy in marketing, we reformulate the CE computation problem as a
profit-maximization problem and propose a framework called profit-based
counterfactual explanation (PBCE).

This study makes three contributions. First, we avoid the exogenous
specification of target values in the CE by introducing a framework in which
profit maximization is treated as the primary objective. That is, the
strategic variables are adjusted to maximize profits, eliminating the need to
specify the external target of the dependent variable. Second, we address the
arbitrariness of the distance function by interpreting it as the cost
associated with changes in product attributes. This interpretation provides an
economically meaningful representation of the mathematical distance measures
commonly used in CE methods and enables the feasibility of counterfactual
solutions to be evaluated from an economic perspective. Third, we conduct
theoretical and empirical analyses to validate the proposed approach.
Specifically, we derive analytical solutions through theoretical analysis,
perform numerical validation through simulations, and conduct an empirical
analysis using data from the Japanese manga market. These analyses demonstrate
the validity of the proposed method and provide practical managerial
implications.

Tsirtsis and G\'{o}mez-Rodr\'{i}guez (2020) formulated CE as a strategic
interaction between a decision-maker and an explainee and proposed a
theoretical model that simultaneously optimizes explanation strategies and
policies \cite{c8}. In their framework, the distance function was interpreted as
effort cost. By contrast, this study reformulates CE within a
profit-maximization framework in management and marketing, explicitly
providing an economic interpretation of both distance and cost. The remainder
of this paper is organized as follows: Section II presents the formulation of
the problem. Section III provides a theoretical analysis. Section IV describes
the proposed method and the empirical analysis. Section V discusses the
results and outlines the directions for future research.

\section{PROBLEM SETTING}

In contrast to conventional CE methods that require an exogenously specified
target value, PBCE reformulates CE as a profit-maximization problem under the
following general setting.

\begin{equation*}
(x_{n}^{*}, p_{n}^{*}) \in
\underset{x_{n},\, p_{n}}{\operatorname{argmax}}\;
p_{n} f(x_{n}, p_{n})
- c\!\left(f(x_{n}, p_{n})\right)
- c_{cg}(x_{n}, x_{n}^{b}).
\end{equation*}

Let $x_{n} = (x_{n,1},\ldots,x_{n,K}) \in \mathbb{R}^{K}$
$(n = 1,2,\ldots,N)$ denote the product attributes of product $n$, where $N$
is the number of products and $K$ is the number of product attributes. Let
$x_{n}^{b} = (x_{n,1}^{b},\ldots,x_{n,K}^{b}) \in \mathbb{R}^{K}$ denote
the baseline $K$-dimensional product attribute vector. $x_{n}$ is a
counterfactual version of the baseline attribute vector $x_{n}^{b}$. The
variable $p_{n} \in \mathbb{R}^{+}$ denotes the price, a one-dimensional
scalar. We assume a monopolistic market structure that serves as a natural
starting point for a profit-based CE analysis.

Function $f$ predicts the quantity demanded $y_{n}$ based on price $p_{n}$
and product attributes $x_{n}$, and is estimated using machine learning
methods. Specifically, $f$ is trained on dataset
$D = \{Y, X, P\}$, which comprises $N$ one-dimensional demand observations
$Y = \{y_{1},y_{2},\ldots,y_{N}\}$, $N$ $K$-dimensional product attribute
vectors $X = \{x_{1},x_{2},\ldots,x_{N}\}$, and $N$ one-dimensional prices
$P = \{p_{1},p_{2},\ldots,p_{N}\}$. We assume that $f(x_{n},p_{n})$
decreases monotonically in $p_{n}$, that is,
$f_{p_{n}}(x_{n},p_{n}) < 0$. Although exceptions may exist in practice, this assumption is motivated by
theoretical consistency and well-established empirical findings on the law of
demand. One possible exception is that a higher price may signal higher quality
and stimulate demand. In the manga market, however, quality signals are primarily
conveyed through reputation, serialization venue, and reader reviews rather
than price, and manga volumes are priced within a narrow, standardized range;
the signaling effect is therefore unlikely to be dominant, making the
monotonicity assumption a reasonable approximation in this context.

Function $c$ denotes the production cost as a function of demand
$f(x_{n},p_{n})$, and is assumed to be monotonically increasing, i.e.,
$c'(\cdot) > 0$. The function $c_{cg}$ represents the adjustment cost for
changing product attributes from the baseline $x_{n}^{b}$ to $x_{n}$, and
can be specified using a distance function. For example, the sum of the
squared differences, or more generally, the weighted Minkowski distance, can
be employed. Such formulations are equivalent to incorporating
attribute-specific adjustment costs $c_{k} \geq 0$, where the non-negativity
ensures that the modification cost is economically well-defined. The two
canonical specifications are as follows.
\begin{align*}
c_{cg}(x_{n}, x_{n}^{b}) &= \sum_{k=1}^{K} c_{k} (x_{n,k} - x_{n,k}^{b})^{2}, \tag{S1} \\
c_{cg}(x_{n}, x_{n}^{b})
&= \left( \sum_{k=1}^{K} c_{k} |x_{n,k} - x_{n,k}^{b}|^{q}
\right)^{\!1/q}. \tag{S2}
\end{align*}

This study considers a setting in which a monopolistic firm sells its products
through online stores or similar channels. In such an environment, the cost of
price adjustment is substantially lower than that of modifying other product
attributes. Accordingly, the adjustment cost function $c_{cg}$ incorporates
only non-price attributes and excludes price from the change cost.

\section{THEORETICAL ANALYSIS}

In this section, we present the theoretical analyses. For simplicity, we
assume that product attribute $x_{n}$ is one-dimensional to gain analytical
tractability while preserving the essential economic structure. We derive the
solution under different model specifications and characterize its key
properties. To simplify the notation throughout this section, we write
$x_{n}$ as $x$, $p_{n}$ as $p$, and $x_{n}^{b}$ as $x^{b}$.

\subsection{General Model}

Let the product attribute be $x \in \mathbb{R}_{+}$, the price be
$p \in \mathbb{R}_{+}$, and the demand quantity be $Q \in \mathbb{R}_{+}$.
Let $f(x,p)$ denote the demand function (predictive model) as defined in
Section II, $c(f(x,p))$ denote the production cost as a function of output
(cf.\ Section II), and $c_{cg}(x, x^{b})$ denote the adjustment cost based
on the distance from the baseline attribute $x^{b} \in \mathbb{R}_{+}$.

We define the firm's profit function as
\[
\Pi(x,p) = p\, f(x,p) - c(f(x,p)) - c_{cg}(x, x^{b}).
\]

\noindent
If the profit function $\Pi$ is strictly concave on $X \times P$, the
solution $(x^{*}, p^{*})$ to the maximization problem is unique. A sufficient
condition for uniqueness is that the stationary point $(x^{*}, p^{*})$ is
unique and the Hessian matrix $H$ is negative definite on $X \times P$.

Accordingly, in addition to satisfying the first-order conditions,
\[
\left\{
\begin{array}{l}
\Pi_{x} = [p - c'(f)]\, f_{x} - c_{cg,x} = 0, \\[4pt]
\Pi_{p} = f + [p - c'(f)]\, f_{p} = 0,
\end{array}
\right.
\]
it is sufficient for the Hessian matrix $H$ to satisfy the second-order
condition of negative definiteness, where
\[
H = \nabla^{2}\Pi(x,p)
= \begin{pmatrix} \Pi_{xx} & \Pi_{xp} \\ \Pi_{px} & \Pi_{pp} \end{pmatrix}.
\]

Here, the second-order partial derivatives are given by
$\Pi_{xx} = [p - c'(f)]\, f_{xx} - c''(f)(f_{x})^{2} - c_{cg,xx}$,
$\Pi_{px} = \Pi_{xp} = (1 - c''(f)f_{p})\,f_{x}
  + [p - c'(f)]\, f_{xp}$,
and
$\Pi_{pp} = 2f_{p} + [p - c'(f)]\, f_{pp} - c''(f)(f_{p})^{2}$.
In this case, the following conditions hold:
\[
\left\{
\begin{array}{l}
\Pi_{xx} < 0, \\[4pt]
\det(H) = \Pi_{xx}\Pi_{pp} - \Pi_{px}\Pi_{xp} > 0.
\end{array}
\right.
\]

Even when $x$ is $K$-dimensional, the maximization problem can be solved
similarly if the corresponding Hessian matrix is negative definite.

\textit{Remark 1 (Marketing Interpretations).} First, consider price optimality given by $\Pi_{p} = 0$.
$c'(f)$ represents the marginal production cost when output increases by one
unit. Accordingly, $[p - c'(f)]$ denotes the price-cost markup. The optimal
price equates the marginal revenue gain from a unit price increase, captured
by $f$, to the marginal revenue loss from the induced reduction in demand,
represented by $[p - c'(f)]\, f_{p}$. Second, consider attribute optimality
given by $\Pi_{x} = 0$. $[p - c'(f)]\, f_{x}$ represents the marginal
revenue from a unit increase in the product attribute, while $c_{cg,x}$
denotes the corresponding marginal adjustment cost. Therefore, the optimal
attribute level is determined by equating marginal revenue with marginal
adjustment cost.

\subsection{Linear Model}

Next, we derive an optimal solution using a linear model specification.
Specifically, we assume $Q = f(x,p) = \alpha + \beta x - \gamma p$,
$c(f(x,p)) = \delta\, f(x,p)$, and
$c_{cg}(x) = \eta(x - x^{b})^{2}$ (based on S1). Without loss of generality, we impose
$\alpha/\gamma > \delta \geq 0$, $\gamma > 0$, $\beta > 0$, $\eta > 0$. To
ensure that the Hessian matrix is negative definite, we additionally assume
$\det(H) = 4\gamma\eta - \beta^{2} > 0$. Under these assumptions, the profit
function is given by
\[
\Pi(x,p) = p(\alpha + \beta x - \gamma p)
           - \delta(\alpha + \beta x - \gamma p)
           - \eta(x - x^{b})^{2}.
\]

Setting $\Pi_{p}=0$ gives $p(x)=(\alpha+\beta x+\gamma\delta)/(2\gamma)$,
and $\Pi_{x}=0$ gives $x(p)=x^{b}+\beta(p-\delta)/(2\eta)$; substituting
these into each other yields the solution below.

Solving this maximization problem yields
\[
p^{*} = \delta + \frac{2\eta(\alpha + \beta x^{b} - \gamma\delta)}
{4\gamma\eta - \beta^{2}},\quad
x^{*} = x^{b} + \frac{\beta(\alpha + \beta x^{b} - \gamma\delta)}
{4\gamma\eta - \beta^{2}}.
\]

In this case, the optimal quantity is
\[
Q^{*} \equiv Q(p^{*})
  = \frac{2\gamma\eta(\alpha + \beta x^{b} - \gamma\delta)}
         {4\gamma\eta - \beta^{2}},
\]
and the optimal profit is
\[
\Pi^{*} \equiv \Pi(x^{*}, p^{*})
  = \frac{\eta(\alpha + \beta x^{b} - \gamma\delta)^{2}}
         {4\gamma\eta - \beta^{2}}.
\]

These closed-form expressions show that the optimal price $p^{*}$ is obtained
by adding $2\eta(\alpha + \beta x^{b} - \gamma\delta)/(4\gamma\eta - \beta^{2})$
to the marginal cost $\delta$, whereas the optimal attribute $x^{*}$ is obtained
by adding $\beta(\alpha + \beta x^{b} - \gamma\delta)/(4\gamma\eta - \beta^{2})$
to the baseline $x^{b}$. Both adjustments share the common factor
$(\alpha + \beta x^{b} - \gamma\delta)/(4\gamma\eta - \beta^{2})$, whose
magnitude is governed by the model parameters.

\section{EMPIRICAL ANALYSIS}

As discussed in Section III, when a simple linear model is assumed, an
optimal solution can be derived analytically. However, obtaining a closed-form
solution is difficult for general nonlinear models or models with interaction
effects. Therefore, we propose a method to estimate a black-box nonlinear
machine-learning model from the observed data and derive the corresponding
PBCE.

\subsection{Empirical Methods}

The proposed method proceeds in the following two steps.

\begin{enumerate}[(i)]
\item \textit{Estimation and comparison of $f$:} We first estimate the demand
  function $f$ using the dataset $D = \{Y, X, P\}$. To do so, monotonicity
  has to be imposed on the price variable $p_{n}$. One possible approach is
  to estimate $f$ using standard regression methods. However, such approaches
  may be insufficient when complex nonlinear relationships and interaction
  effects are present among variables. Therefore, we employ machine learning
  models that enable explicit monotonicity constraints, such as constrained
  monotonic neural networks (CMNNs) \cite{c9}. The predictive performance of the
  estimated models is evaluated and compared using metrics for continuous
  outcomes, such as mean squared error (MSE).

\item \textit{Derivation of optimal $x_{n}$ and $p_{n}$:} Given a baseline
  product attribute vector $x_{n}^{b}$, we derive the optimal price and
  product attributes by solving a constrained nonlinear optimization problem.
  Standard numerical optimization algorithms, such as sequential
  least-squares programming (SLSQP) and trust-region constrained methods, can
  be employed for this purpose. When analytical partial derivatives are
  available, they are used to improve computational efficiency; otherwise,
  numerical differentiation is applied. The optimization problem may admit
  multiple local optima; hence, the procedure is repeated with multiple
  initial values, and the solution that yields the highest profit is selected.
  Deriving CEs for the different values of $x_{n}^{b}$ and comparing their
  distributions and summary statistics enable the identification of variables
  that are consistently important across multiple cases.
\end{enumerate}

\subsection{Evaluation Methods}

In the CE literature, various evaluation metrics have been proposed, including
\textit{Validity}, which measures how close a counterfactual outcome is to a
predefined target value, and \textit{Proximity}, which captures the distance
between an original instance and its CE \cite{c2}. However, the proposed approach is
not intended for evaluation solely in terms of target-oriented validity.
Accordingly, for the PBCE, we introduce the following evaluation metrics and
assess the proposed method based on them.

\textit{Profit:} We evaluate PBCE using the achieved profit $\pi_{n}$ and its
average across instances. Higher values of the optimal profit $\pi_{n}$
indicate better performance. Additionally, we report the corresponding optimal
product attributes $x_{n}^{*}$, prices $p_{n}^{*}$, and predicted demand
$y_{n}^{*}$ as reference values.

\textit{Dissimilarity:} Dissimilarity operationalizes Proximity as the squared Euclidean distance of the changes in non-price attributes, $\mathit{diss} = \sum_{k}(x_{n,k}^{*} - x_{n,k}^{b})^{2}$, reported without the cost coefficient and consistent with $c_{cg}$; smaller values indicate more desirable (lower-cost) CEs.
As reference information, we also examined the magnitude and direction
of changes in the decision variables, including $x_{n}^{*} - x_{n}^{b}$ and
$p_{n}^{*} - p_{n}^{b}$.

\textit{Mean Absolute Error (MAE):} In settings such as simulation studies,
where the optimal values of $x_{n}^{*}$, $p_{n}^{*}$, and $\pi_{n}^{*}$ can
be computed analytically, we evaluate the accuracy of the proposed numerical
optimization method using the mean absolute error (MAE). This evaluation
enables the assessment of the closeness of the numerical solutions to the
theoretical optima. Specifically, we define
$\mathit{diff\_opt}\;x = |\bar{x}_{n}^{*} - x_{n}^{*}|$,
$\mathit{diff\_opt}\;p = |\bar{p}_{n}^{*} - p_{n}^{*}|$,
and $\mathit{diff\_opt}\;\pi = |\bar{\pi}_{n}^{*} - \pi_{n}^{*}|$,
where the overbar denotes the theoretical optimal values.

\subsection{Experiment}

We validate the method proposed in Section IV-A in two stages: a simulation
study and a real-data application. In the simulations, we assess whether the
method could recover the theoretical optimum. We then apply it to real-world
data.

\subsubsection{Simulation Data}

In this simulation, the explanatory variables $x_{n}$ and the price
$p_{n}\ (\geq 0)$ are generated from a uniform distribution on $[0,1]$. This
normalization places all variables on a common scale, which facilitates the
comparability of the adjustment magnitudes and distance functions in CEs.
Notably, this range is purely a scaling choice for data generation and does
not represent economically feasible domains. Therefore, optimal solutions may
lie outside this interval. The error term $\varepsilon_{n}$ is assumed to
follow a standard normal distribution, $N(0,1)$. Using the model specified
below, we generate the outcome variable $y_{n}$ and construct the dataset
$D = \{Y, X, P\}$. A total of 5,000 samples are generated. We further assume
that the firm does not optimize $x_{n}$ and $p_{n}$ with respect to profits
prior to the simulation.
\[ y_{n} = 3 + 2x_{n} - p_{n} + 0.05\varepsilon_{n}. \]

Using dataset $D$, we estimate the demand functions using both linear
regression (LR) and machine learning (ML) approaches. We then compare the
predictive performances of LR and CMNNs with different hyperparameter settings
(e.g., CMNN(3,2): two hidden layers with three and two units) using MSE
(Table~I). For reference, we also compute the MAE. Predictive performance is evaluated using 5-fold cross-validation (CV)
on the full dataset; the reported MSE and MAE values in Table~I represent the
averages across the five folds. Among the ML models, the architecture with the
lowest CV MSE is selected.

\begin{table}[t]
  \caption{Comparison of Model Predictive Accuracy}
  \label{tab:I}
  \centering\small
  \begin{tabular}{lrr}
    \toprule
    Model & MSE & MAE \\
    \midrule
    Linear regression & 0.003 & 0.040 \\
    CMNN(2,2) & 1.785 & 1.013 \\
    CMNN(3,2) & 0.154 & 0.281 \\
    CMNN(4,2) & 0.421 & 0.439 \\
    CMNN(2,3) & 0.334 & 0.436 \\
    CMNN(3,3) & 0.319 & 0.379 \\
    CMNN(4,3) & 0.072 & 0.213 \\
    CMNN(2,4) & 0.312 & 0.424 \\
    CMNN(3,4) & 0.166 & 0.303 \\
    CMNN(4,4) & 0.511 & 0.463 \\
    \bottomrule
  \end{tabular}
\end{table}

The results show that LR achieved the best predictive performance overall,
whereas among the CMNN models, architectures with four and three hidden layers
yielded the lowest prediction errors.

We next compute CEs. The cost functions are $c(Z) = 0.5Z$ and
$c_{cg}(x_{n}, x_{n}^{b}) = 10(x_{n} - x_{n}^{b})^{2}$, and the CEs were
derived based on these cost settings.
We use 15 random initializations with SLSQP; the maximum number of
iterations (30) serves as the convergence criterion, with the default SLSQP
tolerance otherwise applied. No attribute bounds are imposed.

We compare the distributions of the original estimated variables
$x_{n}$, $p_{n}$, $y_{n}$, and $\pi_{n}$ with those of their counterfactual
counterparts $x_{n}^{*}$, $p_{n}^{*}$, $y_{n}^{*}$, and $\pi_{n}^{*}$. We
also examine the differences from the baseline values, including
$|x_{n}^{*} - x_{n}^{b}|$ and $|p_{n}^{*} - p_{n}^{b}|$, to assess the
magnitude of the changes induced by the counterfactual solutions (Table~II; LR-based; cf.\ Table~III for ML-based).

\begin{table}[t]
  \caption{Summary Statistics of CEs (LR)}
  \label{tab:II}
  \centering\footnotesize
  \resizebox{\columnwidth}{!}{%
  \begin{tabular}{lrrrrrrr}
    \toprule
    & $x_n$ & $p_n$ & $y_n$ & $\pi_n$ & $x_n^*\!-\!x_n^b$ & $p_n^*\!-\!p_n^b$ & diss \\
    \midrule
    count & 50 & 50 & 50 & 50 & 50 & 50 & 50 \\
    mean  & 0.424 & 0.424 & 3.302 & 0.069 & 0.186 & 1.813 & 0.036 \\
    std   & 0.279 & 0.279 & 0.639 & 0.910 & 0.031 & 0.426 & 0.012 \\
    min   & 0.001 & 0.001 & 2.088 & $-$1.941 & 0.139 & 0.946 & 0.019 \\
    max   & 0.978 & 0.978 & 4.782 & 1.818 & 0.248 & 2.836 & 0.061 \\
    \bottomrule
  \end{tabular}}
  \par\vskip 0.8em
  \begin{tabular}{lrrrr}
    \toprule
    & $x_n^*$ & $p_n^*$ & $y_n^*$ & $\pi_n^*$ \\
    \midrule
    count & 50 & 50 & 50 & 50 \\
    mean  & 0.610 & 2.359 & 1.860 & 3.197 \\
    std   & 0.310 & 0.310 & 0.310 & 1.060 \\
    min   & 0.140 & 1.888 & 1.389 & 1.735 \\
    max   & 1.225 & 2.974 & 2.476 & 5.513 \\
    \bottomrule
  \end{tabular}
\end{table}

The PBCE achieves profit improvements for many products with relatively limited
attribute adjustments. The average squared attribute change ($\mathit{diss} = 0.036$) is small,
suggesting that profit gains can be realized
without large modifications (mean change in $x$: 0.186). Profit improvements are achieved
primarily through price adjustments, and profit $\pi_{n}$ can vary
considerably. In many cases, the counterfactual profit satisfies
$\pi_{n}^{*} \geq \pi_{n}$, with the maximum counterfactual profit reaching
$\pi_{n}^{*} = 5.51$. Moreover, the results indicate a structure in which
profits increase even when sales volumes decrease. In many cases, $y_{n}^{*}$
is lower than the original value. This finding implies that firms that do not apply this CE-based optimization 
tend to overproduce their quantities. Instead, profit maximization can be
achieved by reducing sales volumes and raising prices, thereby increasing the
margin per unit. From a marketing perspective, this indicates that profit
maximization does not necessarily require sales maximization. Although some
firms pursue sales volume or market share maximization as their primary
objective (e.g., \cite{c10}), the present results suggest that such strategies may
not always be consistent with profit maximization. We discuss this point
further in Section~V.

\begin{table}[t]
  \caption{Summary Statistics of CEs (ML: CMNN)}
  \label{tab:III}
  \centering\footnotesize
  \resizebox{\columnwidth}{!}{%
  \begin{tabular}{lrrrrrrr}
    \toprule
    & $x_n$ & $p_n$ & $y_n$ & $\pi_n$ & $x_n^*\!-\!x_n^b$ & $p_n^*\!-\!p_n^b$ & diss \\
    \midrule
    count & 50 & 50 & 50 & 50 & 50 & 50 & 50 \\
    mean  & 0.424 & 0.424 & 3.303 & 0.081 & 0.207 & 1.476 & 0.066 \\
    std   & 0.279 & 0.279 & 0.710 & 0.911 & 0.154 & 0.632 & 0.081 \\
    min   & 0.001 & 0.001 & 1.840 & $-$1.899 & $-$0.197 & 0.600 & 0.000 \\
    max   & 0.978 & 0.978 & 4.679 & 1.922 & 0.660 & 2.990 & 0.436 \\
    \bottomrule
  \end{tabular}}
  \par\vskip 0.8em
  \begin{tabular}{lrrrr}
    \toprule
    & $x_n^*$ & $p_n^*$ & $y_n^*$ & $\pi_n^*$ \\
    \midrule
    count & 50 & 50 & 50 & 50 \\
    mean  & 0.631 & 2.022 & 2.302 & 2.798 \\
    std   & 0.331 & 0.717 & 0.717 & 1.303 \\
    min   & 0.144 & 1.023 & 0.496 & 1.157 \\
    max   & 1.493 & 3.214 & 3.693 & 6.260 \\
    \bottomrule
  \end{tabular}
\end{table}

Table~III reports the summary statistics of the PBCE derived from the demand
function estimated using the ML model. The average attribute change induced by
CE is moderate (mean = 0.207), indicating that profit improvements require both
price and attribute adjustments. The average price change is 1.476, suggesting
that price remains the most influential decision variable. The average
squared attribute change ($\mathit{diss}$) is small (0.066), indicating that ML-based CEs propose
realistic and feasible levels of change. When we examine the optimal solutions after
applying CE, the minimum profit increases dramatically to 1.157, compared with
$-1.899$ in the original data, implying a substantial reduction in downside
profit risk. The maximum profit reaches 6.260, exceeding the original maximum
profit of 4.679. Although sales decrease on average, an increase in unit price
contributes to higher profits. Compared with the LR-based CEs, the ML-based
CEs exhibit smaller price adjustments (mean = 1.476 vs.\ 1.813). Finally, to assess deviations from the
theoretical optimum, we compare the summary statistics of
$\mathit{diff\_opt}\;x$, $\mathit{diff\_opt}\;p$ and
$\mathit{diff\_opt}\;\pi$, which measure the absolute differences between the
theoretical optimal values and the corresponding numerically obtained solutions
(Table~IV).

\begin{table}[t]
  \caption{Summary Statistics of Differences Between Theoretical and Numerical
           Optimal CE Solutions}
  \label{tab:IV}
  \centering\footnotesize
  \resizebox{\columnwidth}{!}{%
  \begin{tabular}{lrrrrrrrr}
    \toprule
    & \multicolumn{4}{c}{LR} & \multicolumn{4}{c}{ML} \\
    \cmidrule(lr){2-5}\cmidrule(lr){6-9}
    & diff $p$ & diff $x$ & diff $\pi$ & diff $y$
    & diff $p$ & diff $x$ & diff $\pi$ & diff $y$ \\
    \midrule
    count & 50 & 50 & 50 & 50 & 50 & 50 & 50 & 50 \\
    mean  & 0.0014 & 0.0000 & 0.0028 & 0.0003 & 0.468 & 0.114 & 0.547 & 0.598 \\
    std   & 0.0001 & 0.0000 & 0.0001 & 0.0002 & 0.333 & 0.105 & 0.437 & 0.447 \\
    min   & 0.0012 & 0.0000 & 0.0024 & 0.0000 & 0.007 & 0.006 & 0.029 & 0.004 \\
    max   & 0.0015 & 0.0001 & 0.0028 & 0.0006 & 1.379 & 0.434 & 1.795 & 1.584 \\
    \bottomrule
  \end{tabular}}
\end{table}

The LR model yields near-zero average differences from the theoretical optima
($\overline{\mathit{diff}\_p} = 0.0014$,
$\overline{\mathit{diff}\_x} \approx 0$,
$\overline{\mathit{diff}\_\pi} = 0.0028$),
confirming that the proposed numerical optimization accurately recovers the
analytical solution. By contrast, the ML (CMNN) model exhibits larger
differences ($\overline{\mathit{diff}\_p} = 0.468$,
$\overline{\mathit{diff}\_\pi} = 0.547$). This is expected: the theoretical
optimum is derived from the linear model parameters, whereas the CMNN
numerical solution reflects a different nonlinear model structure.
Accordingly, the ML differences quantify model divergence rather than
optimization error.

Overall, these results demonstrate that the proposed PBCE accurately recovers
the analytical optimum when a correctly specified demand model is used. The
near-zero LR differences confirm the validity of the numerical procedure, while
the larger ML differences reflect inherent model misspecification relative to
the linear theoretical benchmark.

\subsubsection{Real Data}

Next, using data on manga (Japanese comics) sales volumes, prices, and other
attributes in Japan, we examine how each manga title can be modified to achieve
higher profits. We assume that product attributes remain constant over the
publication period of each manga series. For empirical data, we collected the top
206 manga titles in terms of cumulative sales in Japan as of 2023 \cite{c11}.
Two titles were excluded due to missing attribute data, yielding a final
sample of 204 observations.
The sample size is limited by the availability of publicly accessible sales
information, and the dataset is relatively small for machine learning analysis;
therefore, the results should be interpreted with caution because of the
potential risks of overfitting and estimation instability. From these data, we
computed an approximate per-volume sales figure by dividing the cumulative
sales of each manga by the number of volumes published and using this value as
$Y$. This procedure implicitly assumes that sales are uniformly distributed
across volumes. In practice, earlier volumes tend to sell more copies than
later volumes. However, this is a simplifying assumption, providing a
tractable and consistent measure of average commercial performance at the
series level, which is the unit of analysis in this study. Next, based on the
total price of all newly published print volumes and the number of volumes, we
calculate the approximate average price per volume, which is used as $P$.
Finally, as an explanatory variable $X$, we employ the eight attributes listed
below. Manga inherently contains rich information at the episode level and
exhibits complex sequential structures across episodes. However, as quantifying
such information in a consistent and comparable manner is difficult, this study
focused on work-level attributes related to the overall visual elements and
story characteristics of each manga title. This issue will be discussed in the
following section. The eight attributes are as follows.

(1) Number of years since first publication. This variable is constructed based
on the year the manga was first published. (2) Serialization of the manga in
the \textit{Weekly Shonen Jump}. The \textit{Weekly Shonen Jump} is the most
popular weekly manga magazine in Japan; a large share of top-selling manga
titles have been serialized in this magazine. Therefore, we include this
attribute. (3--5) Three feature vectors related to manga's visual
characteristics. For each manga title, five images are collected using an
image search engine, excluding non-manga images. These images are then embedded
using a Vision Transformer (ViT), resulting in 768-dimensional feature vectors
\cite{c12}. We subsequently apply principal component analysis (PCA) and reduce the
dimensionality to three components selected based on scree plots and semantic
interpretability. The resulting visual components are interpreted as follows
(Table~V): \textit{vit1} represents a spectrum from dramatic (high values:
cinematic and intense visual styles) to gentle (low values: everyday and soft
visual styles). \textit{vit2} captures a range from realistic (high values:
greater realism and stronger shading) to cartoonish (low values: deformation
and a simplified style). \textit{vit3} reflects a continuum ranging from a
stylized appearance (high values: mature and fashionable style) to youthful
characteristics (low values: youthfulness and cuteness). (6--8) Three feature
vectors related to the story and building characteristics. Using textual data,
we construct three features based on story descriptions. Specifically, we
extract the English-language summaries of each manga from MyAnimeList \cite{c13}.
These texts are embedded using BERT, yielding 768-dimensional vectors \cite{c14}. We
then apply PCA to these embeddings and compress them into three components that
are selected based on scree plots and semantic interpretability. Based on the
interpretation of high-loading examples, the textual components are defined as
follows (Table~VI): \textit{bert1} represents a spectrum ranging from heroic
(high values: battles, death, fate, and extraordinary events) to mundane (low
values: everyday life, humor, and family). \textit{bert2} captures a continuum
from masculine (high values: social roles and occupations) to emotional (low
values: emotions and aesthetics). \textit{bert3} reflects a range from epic
(high values: history, fantasy, and mythology) to contemporary (low values:
modern settings, youth, and school life). PCA is conducted independently for
the visual and textual features. As these features belong to different
representational spaces, their variance structures are not directly comparable,
reflecting the differences in the amount and nature of information captured by
each modality.

\begin{table}[t]
  \caption{Top and Bottom Values of Feature Dimensions (ViT)}
  \label{tab:V}
  \centering\footnotesize
  \resizebox{\columnwidth}{!}{%
  \begin{tabular}{lrlrlr}
    \toprule
    Title & vit1 & Title & vit2 & Title & vit3 \\
    \midrule
    Fist of the North Star   &  1.948 & Record of Ragnarok   &  1.009 & Coffee \& Vanilla            &  1.723 \\
    Toriko                   &  1.644 & Black Butler         &  0.981 & Initial D                    &  1.484 \\
    Eyeshield 21             &  1.642 & Yamikin Ushijima-kun &  0.972 & The King of Minami           &  1.040 \\
    Dragon Ball              &  1.507 & Terra Formars        &  0.953 & GANTZ                        &  0.989 \\
    Kinnikuman               &  1.471 & Vagabond             &  0.933 & BLEACH                       &  0.963 \\
    \midrule
    Video Girl Ai            & $-$1.455 & Crayon Shin-chan      & $-$1.131 & To Love-Ru                   & $-$0.880 \\
    Atashin'chi              & $-$1.464 & Kochikame            & $-$1.146 & Honey Lemon Soda             & $-$0.927 \\
    His and Her Circumstances& $-$1.501 & Sazae-san            & $-$1.206 & Ao Haru Ride                 & $-$0.973 \\
    Animal Doctor            & $-$1.525 & High School! Kimengumi& $-$1.305 & The Quintessential Quintuplets& $-$1.001 \\
    Chibi Maruko-chan         & $-$1.556 & Tsuribaka Nisshi     & $-$1.326 & Rurouni Kenshin              & $-$1.017 \\
    \bottomrule
  \end{tabular}}
\end{table}

\begin{table}[t]
  \caption{Top and Bottom Values of Feature Dimensions (BERT)}
  \label{tab:VI}
  \centering\footnotesize
  \resizebox{\columnwidth}{!}{%
  \begin{tabular}{lrlrlr}
    \toprule
    Title & bert1 & Title & bert2 & Title & bert3 \\
    \midrule
    GANTZ                    &  5.608 & The King of Minami      &  5.443 & D.Gray-man                        &  5.030 \\
    Seraph of the End        &  5.029 & Ranma 1/2               &  5.348 & Sukeban Deka                      &  4.308 \\
    Tokyo Ghoul              &  4.880 & Rokudenashi BLUES       &  5.298 & Nausica\"a of the Valley of the Wind&  3.934 \\
    NARUTO                   &  4.857 & Yu-Gi-Oh!               &  5.245 & Romance of the Three Kingdoms     &  3.725 \\
    Maison Ikkoku            &  4.590 & Animal Doctor           &  4.619 & Record of Lodoss War              &  3.701 \\
    \midrule
    YAWARA!                         & $-$6.113 & Urusei Yatsura    & $-$3.713 & Play Ball                         & $-$3.210 \\
    Asari-chan               & $-$6.151 & The Rose of Versailles  & $-$3.736 & Silver Spoon                      & $-$3.319 \\
    Sazae-san                & $-$6.393 & Tenjho Tenge            & $-$4.138 & Ao Haru Ride                      & $-$3.405 \\
    Osomatsu-kun             & $-$6.496 & ONE PIECE               & $-$4.236 & H2                                & $-$3.426 \\
    Kaguya-sama: Love Is War & $-$6.722 & Record of Ragnarok      & $-$4.354 & Itazura na Kiss                   & $-$3.468 \\
    \bottomrule
  \end{tabular}}
\end{table}

Tables~VII and~VIII report the summary statistics for the dependent and
explanatory variables, respectively.

\begin{table}[t]
  \caption{Summary Statistics of the Dependent Variable}
  \label{tab:VII}
  \centering\footnotesize
  \resizebox{\columnwidth}{!}{%
  \begin{tabular}{lrrrrr}
    \toprule
    & circ\_num & volume & price & per\_num & per\_price \\
    \midrule
    count &       204 &     204 &        204 &     204 &      204 \\
    mean  & 4067.775 &  29.711 & 18627.931  & 169.861 &  741.177 \\
    std   & 5468.118 &  29.628 & 15896.025  & 156.789 &  477.643 \\
    min   &    1000   &       1 &      748   &  12.200 &  104.500 \\
    max   &   50000   &     211 &   121508   & 1000.000 & 5280.000 \\
    \bottomrule
  \end{tabular}}
\end{table}

\begin{table}[t]
  \caption{Summary Statistics of the Explanatory Variables}
  \label{tab:VIII}
  \centering\footnotesize
  \resizebox{\columnwidth}{!}{%
  \begin{tabular}{lrrrrrrrr}
    \toprule
    & year & jump & vit1 & vit2 & vit3 & bert1 & bert2 & bert3 \\
    \midrule
    count &    204 & 204 &    204 &    204 &    204 &    204 &    204 &    204 \\
    mean  & 26.279 & 0.265 & $-$0.006 &  0.009 & $-$0.004 &  0.000 &  0.000 &  0.000 \\
    std   & 14.670 & 0.442 &  0.791 &  0.505 &  0.457 &  2.898 &  2.140 &  1.747 \\
    min   &      4 &     0 & $-$1.556 & $-$1.326 & $-$1.017 & $-$6.722 & $-$4.354 & $-$3.468 \\
    max   &     78 &     1 &  1.948 &  1.009 &  1.723 &  5.608 &  5.443 &  5.030 \\
    \bottomrule
  \end{tabular}}
\end{table}

Predictive accuracy is evaluated using 5-fold CV. Prior to model
fitting, all explanatory variables are standardized to zero mean and unit
variance for model input only; the optimization and cost function operate on
the original scale. CMNN training uses a batch size of 16, a maximum of 100 epochs, and
early stopping (patience of 5 epochs) to prevent overfitting. The CMNN with
the lowest CV MAE is selected as the base model for CE derivation.
Table~IX reports the predictive accuracy of the LR and several CMNN models.
According to the MSE and MAE metrics, the LR exhibits the smallest errors,
indicating a strong fit between the linear model and the data. Among the CMNN
models, CMNN(5,5,5) achieves the best performance (MSE = 26382.287, MAE =
94.791), with error levels close to those of LR. This result suggests that the
relationship between the variables is approximately linear, although the
relatively small sample size ($N = 204$) may also limit the performance of
more flexible machine learning models. However, the CMNN models achieve
comparable predictive accuracies.

\begin{table}[t]
  \caption{Comparison of Model Predictive Accuracy}
  \label{tab:IX}
  \centering\small
  \begin{tabular}{lrr}
    \toprule
    Model & MSE & MAE \\
    \midrule
    Linear regression & 23502.243 &  96.080 \\
    CMNN(4,5,5)       & 31885.983 & 109.645 \\
    CMNN(4,5,6)       & 26368.902 &  96.237 \\
    CMNN(4,6,5)       & 26449.038 &  98.928 \\
    CMNN(5,5,5)       & 26382.287 &  94.791 \\
    CMNN(5,5,4)       & 27069.018 &  96.337 \\
    \bottomrule
  \end{tabular}
\end{table}

In the context of the PBCE, employing models that satisfy economic constraints,
such as a monotonic decrease in demand with respect to price, is particularly
important. CMNNs can directly incorporate the monotonicity constraints.
Although their prediction errors are slightly larger than those of LR, CMNNs
offer the advantage of balancing economic consistency and model flexibility.
Therefore, in this study, we adopt the CMNN(5,5,5) as the base model for CE,
as it preserves theoretical consistency while maintaining a sufficiently high
predictive accuracy.

We now derive the CEs. For simplicity, the cost functions are
specified as $c(Z) = 10Z$ and $c_{cg}$ based on S1
with a common coefficient $c_k = 500$ for all $k$, giving
$c_{cg}(x_{n}, x_{n}^{b}) = 500\sum_{k=1}^{K}(x_{n,k} - x_{n,k}^{b})^{2}$.
We use 5 random initializations with SLSQP; the maximum number of
iterations (30) serves as the convergence criterion, with the default SLSQP
tolerance otherwise applied. \textit{year} and \textit{jump} are held fixed
via equality constraints; PCA-based features are bounded to $[-3, 3]$; and
price is left effectively unconstrained. We now turn to the results of the
CEs (Table~X). In Tables~X and XI, \textit{base} denotes the baseline
value, \textit{cf} the counterfactual value, \textit{diff} the difference
(cf$-$base), and \textit{demand} the predicted demand.

\begin{table}[t]
  \caption{Counterfactual Explanations for a Representative Case}
  \label{tab:X}
  \centering\footnotesize
  \resizebox{\columnwidth}{!}{%
  \begin{tabular}{llrrrrrrr}
    \toprule
    & & price & year & jump & vit1 & vit2 & vit3 \\
    \midrule
    \multirow{3}{*}{NARUTO}
      & base & 484.300 & 25 & 1 &  0.977 & $-$0.675 &  0.143 \\
      & cf   & 484.296 & 25 & 1 &  0.903 & $-$0.518 &  0.174 \\
      & diff & $-$0.004 & 0 & 0 & $-$0.074 &  0.157 &  0.031 \\
    \midrule
    \multirow{3}{*}{Doraemon}
      & base & 567.100 & 55 & 0 & $-$1.301 & $-$0.937 & $-$0.101 \\
      & cf   & 567.208 & 55 & 0 & $-$1.232 & $-$0.862 & $-$0.122 \\
      & diff &   0.108 &  0 & 0 &  0.069 &  0.075 & $-$0.020 \\
    \bottomrule
  \end{tabular}}
  \par\vskip 0.8em
  \resizebox{\columnwidth}{!}{%
  \begin{tabular}{llrrrrrr}
    \toprule
    & & bert1 & bert2 & bert3 & demand & profit & diss \\
    \midrule
    \multirow{3}{*}{NARUTO}
      & base &  4.857 &  3.519 &  0.733 & 158.700 & 75271.305 & \\
      & cf   &  4.894 &  3.493 &  0.856 & 162.040 & 76830.617 & \\
      & diff &  0.037 & $-$0.027 &  0.123 &         &           & 0.048 \\
    \midrule
    \multirow{3}{*}{Doraemon}
      & base & $-$3.269 &  2.837 & $-$0.759 & 120.898 & 67352.297 & \\
      & cf   & $-$3.261 &  2.961 & $-$0.954 & 123.424 & 68740.805 & \\
      & diff &  0.008 &  0.124 & $-$0.195 &         &           & 0.064 \\
    \bottomrule
  \end{tabular}}
\end{table}

\textit{Case 1: Naruto.}
For Naruto, CE shifts the visual style toward realism and strong
shading (\textit{vit2}) and strengthens epic elements (\textit{bert3}),
increasing both demand and profit. In practice, the \textit{vit2} shift
suggests enhancing shading and anatomical detail---adjustments feasible at the
character-design and panel-composition stages.

\textit{Case 2: Doraemon.}
For Doraemon, CE adjusts visual features toward more dramatic and
realistic styles (\textit{vit1}, \textit{vit2}) and shifts the narrative
toward social and contemporary themes (\textit{bert2}, \textit{bert3}), also
modifying price. Compared with Naruto, Doraemon requires adjustments across a
larger number of elements. In practice, the \textit{bert2} and \textit{bert3}
shifts suggest foregrounding everyday social interactions and contemporary
settings---achievable through scenario selection.

Across all titles, ViT-based features shift toward greater realism
(\textit{vit2}) and stylized designs (\textit{vit3}), while BERT-based
features strengthen heroic (\textit{bert1}) and epic (\textit{bert3}) themes
(Table~XI). Average demand increases and profit improves from 94,612 to
97,603, confirming that PBCEs identify profit-improving solutions through
content-related adjustments rather than exogenous conditions such as price or
publication information.

\begin{table}[t]
  \caption{Average Counterfactual Adjustments Across All Titles}
  \label{tab:XI}
  \centering\footnotesize
  \resizebox{\columnwidth}{!}{%
  \begin{tabular}{llrrrrrrr}
    \toprule
    & & price & year & jump & vit1 & vit2 & vit3 \\
    \midrule
    \multirow{3}{*}{Total (avg)}
      & base & 741.177 & 26.279 & 0.265 & $-$0.006 &  0.009 & $-$0.004 \\
      & cf   & 741.177 & 26.279 & 0.265 & $-$0.013 & 0.031 & 0.012 \\
      & diff &   0.000 &  0.000 & 0.000 & $-$0.007 & 0.022 & 0.016 \\
    \bottomrule
  \end{tabular}}
  \par\vskip 0.8em
  \resizebox{\columnwidth}{!}{%
  \begin{tabular}{llrrrrrr}
    \toprule
    & & bert1 & bert2 & bert3 & demand & profit & diss \\
    \midrule
    \multirow{3}{*}{Total (avg)}
      & base & 0.000 &  0.000 & 0.000 & 128.251 & 94612.352 & \\
      & cf   & 0.032 & $-$0.015 & 0.009 & 132.232 & 97602.943 & \\
      & diff & 0.032 & $-$0.015 & 0.009 &         &           & 0.078 \\
    \bottomrule
  \end{tabular}}
\end{table}

\section{DISCUSSION}

We propose a CE framework that reformulates profit maximization as its primary
objective, with a method for deriving such explanations and tailored evaluation
metrics. Theoretical analysis provides analytical solutions characterizing the
key properties of PBCE. Simulation results confirm that the method recovers
solutions close to the theoretical optima, and real-data analysis yields
practically meaningful managerial insights.

In the literature on CEs and algorithmic recourse, the dominant approach is to
generate counterfactuals that achieve the desired output, while distances are
either exogenously specified or interpreted as effort or adjustment costs
\cite{c4,c8}. Moreover, when regression or continuous outcomes are considered, the
choice of target value is often determined exogenously (e.g., \cite{c2}). By
contrast, our approach directly maximizes profit as the primary managerial
objective, thereby avoiding the problem of target specification. Furthermore,
by interpreting distances as costs associated with changes in products, prices,
or managerial actions, the proposed profit-based CE provides explanations based
on the practical constraints of managerial decision making. This finding
resonates with the long-standing debate in marketing that profit maximization
does not necessarily require sales maximization \cite{c10,c15}. The results provide a
data-driven illustration of this divergence, suggesting that firms oriented
toward sales or market share maximization may be forgoing substantial profit
gains.

This study has several limitations. First, there is the issue of the
robustness of CEs \cite{c16}, namely, how robust the derived CEs are to model
misspecification, estimation errors, or changes in the data-generating process.
In this study, CEs were extracted from a single predictive model; however,
more robust explanations can be obtained by deriving CEs from multiple models
and aggregating the results. Second, the models employed do not guarantee causal relationships; PBCEs are
derived from predictive models and do not directly represent intervention
effects. Although predictive models may capture certain causal patterns
\cite{c17}, explicit causal modeling would further improve the reliability of
recommendations. Third, there is an issue of data bias.
The manga data analyzed in this study include many popular titles and are
concentrated in specific media outlets (e.g., serialization in
\textit{Weekly Shonen Jump}). Therefore, caution should be exercised when
generalizing these findings to other media platforms or genres. In addition,
visual and textual features rely on data obtained from search engines and
MyAnimeList, making it difficult to fully eliminate potential information bias.

Finally, we conclude by discussing possible extensions and future research
directions. First, the proposed method applies to a wide range of goods (e.g., consumer
products); more rigorous estimation of adjustment costs from historical data or
expert judgment \cite{c18} would assess the feasibility of recommended changes.
A further extension would learn cost functions from decision-making histories. Second, in this
study, no distance or adjustment cost was imposed on price $p_{n}$, based on
the assumption that price changes are relatively easy to implement,
particularly in online settings (e.g., \cite{c19}). However, in contexts such as
printed books, where prices are printed directly on products, or in
brick-and-mortar retail environments where price changes incur operational
costs, it is necessary to incorporate price adjustment costs. This can be
addressed by introducing a cost function of the form
$c_{cg}\!\left((x_{n},p_{n}),(x_{n}^{b},p_{n}^{b})\right)$ that explicitly
incorporates price changes. Third, the cost functions employed---linear
production costs and quadratic attribute adjustment costs---represent one
particular parametric specification. Alternatives such as economies of scale in
production or heterogeneous attribute-specific adjustment costs would alter
PBCE solutions. Even under nonlinear production costs such as an S-shaped cost curve,
the qualitative result that a monopolist equates marginal revenue (MR) with
marginal cost (MC) is preserved. Heterogeneous attribute adjustment costs would concentrate recommended
changes on lower-cost attributes, yielding more targeted counterfactuals;
examining such sensitivity is an important direction for future research.
Finally, while the present analysis treats each manga title as a single unit,
future work could extend the framework to episode-level analysis using text
analysis and image features. Extending the framework to oligopolistic
or competitive market structures constitutes another promising direction. Such
an extension requires modifying the demand function to incorporate rivals'
prices and attributes, with Nash equilibrium replacing the monopoly optimum as
the solution concept. Competition would generally be expected to lower
equilibrium prices and increase the distance between a firm's own product
attributes and those of its rivals relative to the monopoly benchmark,
while the specific direction of product-attribute adjustments would depend
on the oligopoly model structure, altering both the direction and magnitude
of PBCE-recommended adjustments.

\addtolength{\textheight}{-8cm}   

\section*{ACKNOWLEDGMENT}

This study was supported by the JSPS KAKENHI Grants-in-Aid for Scientific
Research (C) JP21K01468, JP25K05082, and JP25K05381.


\end{document}